\documentclass[conference]{IEEEtran}
\IEEEoverridecommandlockouts
\usepackage{cite}
\usepackage{amsmath,amssymb,amsfonts}
\usepackage{algorithmic}
\usepackage{graphicx}
\usepackage{textcomp}
\usepackage{xcolor}
\usepackage[numbers]{natbib}
\usepackage{wrapfig}

\def\BibTeX{{\rm B\kern-.05em{\sc i\kern-.025em b}\kern-.08em
    T\kern-.1667em\lower.7ex\hbox{E}\kern-.125emX}}
\begin{document}

\title{From Interactive to Co-Constructive Task Learning
%Workshop paper submission deadline: 28th April 2023 (AOE)
\\
\thanks{This research was funded by the Deutsche Forschungsgemeinschaft (DFG, German Research Foundation): TRR 318/1 2021 – 438445824 ``Constructing Explainability'' and SFB 1320 - 329551904 ``EASE - Everyday Activity Science and Engineering''.}
}

\author{\IEEEauthorblockN{1\textsuperscript{st} Anna-Lisa Vollmer}
\IEEEauthorblockA{\textit{Medical Assistance Systems} \\
\textit{CITEC, Medical School EWL}\\
\textit{Bielefeld University}\\
Bielefeld, Germany \\
0000-0002-9378-7249}
\and
\IEEEauthorblockN{2\textsuperscript{nd} Daniel Leidner}
\IEEEauthorblockA{\textit{Fault-tolerant Autonomy Architectures} \\
\textit{Institute of Robotics and Mechatronics} \\
\textit{German Aerospace Center (DLR)}\\
Weßling, Germany \\
0000-0001-5091-7122}
\and
\IEEEauthorblockN{3\textsuperscript{rd} Michael Beetz}
\IEEEauthorblockA{\textit{Institute for Artificial Intelligence (IAI)} \\
\textit{University of Bremen}\\
Bremen, Germany \\
0000-0002-7888-7444}
\and
\IEEEauthorblockN{4\textsuperscript{th} Britta Wrede}
\IEEEauthorblockA{\textit{Medical Assistance Systems} \\
\textit{CITEC, Medical School EWL}\\
\textit{Bielefeld University}\\
Bielefeld, Germany \\
0000-0003-1424-472X}
}

\maketitle

% \begin{abstract}

% \end{abstract}

% \begin{IEEEkeywords}
% Co-construction, Robot Learning, Learning from Demonstration, Interactive Task Learning, robot teaching, scaffolding, developmental robotics
% \end{IEEEkeywords}

\section{Introduction}
With growing robot capabilities, the number of potential fields of employment of robots is also growing. Many of these fields involve the interaction with humans. Robots have for instance been suggested to hold the potential for alleviating the nursing crisis in rapidly aging societies as assistance systems. For the employment of robots in the real world, robots necessarily require the capability to adapt to a-priori unknown environments and to learn the preferences and demands of human users that are laypersons in robotics. For example, a user with a broken arm requires specific help in tasks such as holding a bread that the user wants to cut, but also the user might change her usual preferences due to the temporary impairment. Thus, the encountered environments and demanded tasks are diverse and highly individual, especially in the fields of medicine and care, such that the only viable solution is to learn directly from the human users. From the technical perspective, the fields of imitation learning and learning from demonstration have long developed approaches and machine learning algorithms for learning tasks from humans, for an overview see \citep{argall2009survey}. However, these approaches are rarely evaluated with naive users within an everyday context which may change important parameters of a task.

Learning to adapt tasks to new contexts is thus an important capability for robots - but also an extremely difficult one as most state-of-the-art algorithms require large amounts of data to learn to generalize to different contexts. Humans, in contrast, have developed the capability to teach relevant aspects of new or adapted tasks to a social peer %thereby taking into account the current limits of the learner and what s/he can potentially achieve, i.e. the zone of proximal development.
with very few task demonstrations by making use of scaffolding strategies that leverage prior knowledge and importantly prior joint experience to yield a joint understanding and a joint execution of the required steps to solve the task. 
This process has been discovered and analyzed in parent-infant interaction by \citet{bruner1984vygotsky} and constitutes a ``co-construction'' as it allows both, the teacher and the learner, to jointly contribute to the task.

We propose to focus research on this co-construction process to enable robots to learn from non-expert users in everyday situations.
In the following, we will review current proposals for interactive task learning and discuss their main contributions with respect to the entailing interaction. We then discuss our notion of co-construction and summarize research insights from adult-child and human-robot interactions to elucidate its nature in more detail.
From this overview we finally derive research desiderata that entail the dimensions architecture, representation, interaction and explainability. 

\section{Related work}
\subsection{Interactive Task Learning}
\citet{laird2017interactive} have proposed a new research area they call \emph{Interactive Task Learning (ITL)}. ITL is concerned with how an agent learns an initial representation of a task (i.e., understands its meaning) from which the agent can then learn to do the task well. This initial representation is learned from human instruction in natural interaction.
Thus, \citet{laird2017interactive} seem to mainly be concerned with verbal communication.
%\subsubsection{Notion of Natural Interaction and its Purpose}
According to \citet{laird2017interactive}, an interaction is initiated by the tutor (which the authors refer to as instructor) or learner using language. In this initialisation, an overview of the task is given. During the interaction, the tutor describes the task, including its purpose, when it is appropriate, its constraints, and its termination condition.
Depending on the student's knowledge, the tutor provides feedback, scaffolding, and step-by-step instructions, which occur when the learner is attempting to execute the task, as well as demonstrations in case the task is a physical activity. The learner acquires the meaning of the task while performing it.
In case of ambiguity in the instruction or when the used terminology is unclear, the learner can ask the tutor for clarification.

%\subsubsection{Requirements and Approaches}
\citet{laird2017interactive} argue that the instructions the tutors provide must be interpreted task-independently.
%focus on generality, effectiveness, and efficiency across task learning, performance, and interaction

% \subsection{Interaction for Task Instruction and Learning}

% \subsubsection{Notion of Natural Interaction and its Purpose}

% \subsubsection{Requirements and Approaches}

\subsection{Teachable, Autotelic Agents}
\citet{sigaud2022towards} contrast instructions and autonomous exploration of agents.
Accordingly, teachable agents such as interactive reinforcement learners do not learn autonomously and are missing the ability to set and autonomously achieve their own goals, to learn in an open-ended manner, based on few examples, and to combine and reuse skills for tasks of higher complexity. On the other hand, the authors argue that autotelic agents that are designed to set their own goals and learn them based on their own learning signals \citep{steels2004autotelic} are missing the ability to incorporate external teaching signals from human tutors. The authors point out the potential benefits of a mixed approach.

%\subsubsection{Notion of Natural Interaction and its Purpose}
The notion of natural interaction \citet{sigaud2022towards} put forward is derived from \citep{wood1976role}. They focus on the teacher guiding and assisting the learner in the discovery of a task through interventions. 
%- tutor needs to monitor a model of the knowledge, hypotheses and performance of children as well as a model of the task itself
Interaction, thus, serves assistance in this discovery process. Importantly, concerning the purpose of interactions and tutor input, they put forward that providing demonstrations is in fact a non-verbal way to communicate about intermediate goals rather than about the way to achieve them and that tutoring interactions are in charge of regulating the motivational system (towards the task goal) through recruitment, direction maintenance, and frustration control \citep{wood1976role}, which is proposed to be the role of verbal communication.
Additionally, \citet{sigaud2022towards} state that socio-cultural skills are learned in social interaction which enables the acquisition of shared cognitive representations.
They call for autonomous, hierarchical few-shot learners.

\section{Co-Construction in Task Learning}

The term \emph{co-construction} has its roots in the humanities, social sciences and linguistics. Child language studies have investigated co-construction most extensively.
The definition of the term, however, is not entirely clear. As \citet{jacoby1995co} aptly put it, co-construction denotes the
\begin{quote}
    "fundamentally interactional basis of the human construction of meaning, context, activity, and identity. This is to say (1) that things allegedly in people's heads-such as cognition and attitudes, linguistic competence, or pragmatic and cultural knowledge are made relevant to communication through social interaction, and (2) that it is through the spontaneous playing out of the sequentially contingent and co-constructed external flow of interactional events that human beings bring these conscious, semiconscious, and unconscious internal constructs and potentialities to bear on the constitution, management, and negotiation of social reality and social relationships."
\end{quote}
Interactional events, as the authors stress, do not only pertain to verbal utterances of interactants. They are multimodal and involve for example verbal cues and utterances, eye gaze, facial expressions, gestures and combinations of these as well as the suppression of activity and ensuing pauses. Co-construction spans observable communicative actions and cognitive processes.
In learning and teaching interactions, the co-construction concerns roles, interaction patterns, meaning, utterances, non-verbal behavior, and most importantly interaction goals that are jointly defined in interaction in a process of mutual adaptation.

\subsection{Characteristics of Co-Construction in Adult-Child Interaction}

\citet{wood1976role} emphasize the role of scaffolding in adult-child teaching-learning interactions, the adaptations in teaching behavior in tutors that support the learner at any time in successfully achieving tasks in the Zone of Proximal Development (ZPD). The ZPD was introduced by Vygotsky and describes the set of tasks a learner cannot yet do on their own but is able to achieve with the support from somebody else, for instance a caregiver \citep{bruner1984vygotsky}. We consider scaffolding to be a vital part of co-construction. It has been observed in multimodal communication towards children in speech (`Motherese' \citep{newport1975motherese}) and motion (`Motionese' \citep{brand2002evidence}) and seems to be preferred by children \citep{fernald1985four,brand2008infants}. However, scaffolding is not only a fixed adaptation towards a specific interaction partner (e.g., child vs. adult or children of different age \citep{rohlfing2022motionese}). It is highly adaptive. It has been shown that scaffolding takes place on multiple levels and time scales, and towards multiple goals. On a micro-level, caregivers adjust their action-demonstrations towards scaffolding a child's attention in an interaction loop unfolding moment-by-moment \citep{pitsch2014tutoring}. During interactions, children's behavior has been investigated. Their behavior can naturally be monitored and interpreted by tutors and presents a feedback to tutors about the children's current understanding. Whereas small children's gaze primarily informs tutors continuously about children's attention and understanding of goals (anticipating gaze), older children provide feedback in form of gestures and verbalizations at specific moments in the structure of the task \citep{vollmer2010developing}. When presenting a task to the learner in a demonstration, tutors synchronize their speech with their movements, also called `Acoustic Packaging', to convey the structure of the shown task \citep{brand2007acoustic}. Another means with which structure is conveyed is the suppression of body movements and social signals to highlight the important states of objects such as the initial and the final state \citep{nagai2008toward}. 
Over multiple interactions, we can observe recurring multimodal patterns of behavior between tutor and learner: `Pragmatic Frames' \citep{rohlfing2016alternative}. Pragmatic Frames carry interactional meaning and are presumed to be tied to underlying cognitive processes. They evolve in interaction and are a product of co-construction in which each interaction partner has their role and tasks to fulfill towards a joint goal which is also subject to co-construction. 

% \begin{itemize}
%     \item scaffolding, zone of proximal development (Bruner)
%     \item interaction loop   Karola
%     \item child feedback (anticipatory gaze) Karola
%     \item feedback for children with diff. age
    
%     \item acoustic packaging Lars
%     \item structure through unterdrueckung von bewegung/signalen am TaskEnde   Claudia, Yukie
%     \item Pragmatic Frames
%     \item Ostension - Katrin
%     \item interactional tension ("Spannungsbogen")
%     \item detection of knowledge gaps
% \end{itemize}

% something new is emerging in interaction, in the process of mutual adaptation, that has added value.

% Goals are changing

\subsection{Co-construction in Human-Robot Interaction}
\subsubsection{Computational Benefits of Child-Directed Interaction}
The co-construction and its essential behavior adaptation from teacher to learner (i.e., scaffolding) potentially bears benefits for robot learners. 
There is evidence that in certain cases scaffolding strategies such as hyperarticulation can indeed help current machine learning approaches to learn models that can generalize better \citep{philippsen2017hyperarticulation} even without explicitly extracting the scaffolding cues prior to learning.
Similarly, the scaffolding strategy of synchrony, i.e. presenting speech in synchrony with the demonstrated action parts (Acoustic Packaging) \citep{schillingmann2009computational} helps to detect meaningful elements in the demonstration \citep{wrede2013making} and can be measured on the signal-level \citep{rolf2009attention}. 
Indeed, \citet{saran2020understanding} have shown that it is possible to use teacher gaze patterns to structure a complex task into sub-tasks which are easier to learn.

\subsubsection{Scaffolding Towards Robots}
The above works have mostly investigated benefits of scaffolding on teaching input from adult-child interactions. For human-robot interaction, this input can look differently and the scaffolding modifications do not only rely on a register-based adaptation towards the learner but strongly depend on the learner's feedback in interaction \citep{vollmer2009people}. A simple saliency-based gaze behavior \citep{nagai2008toward} seems to serve as a channel for communication but is not adequate for the same level of tailoring and recipient-oriented scaffolding as in adult-child interaction \citep{vollmer2009people}. The possibilities a robot has on exhibiting feedback seems to influence scaffolding as has been shown for saliency-based gazing and the degrees of freedom of a simulated and embodied robot \citep{lohan2010does}.

What becomes evident in previous research is that human teachers do adapt to the feedback of a robot. \citet{muhl2007does} have shown that humans try to regain the attention of an inattentive robot in interaction. They align their utterances and actions to the ones of a robot for communicative success \citep{iio2009lexical,vollmer2015alignment}. They adapt their scaffolding to a robot on a micro-level during demonstrations in interactive loops \citep{pitsch2013robot} and also on a broader scale from demonstration to demonstration based on the robot's performance \citep{vollmer2013demonstrating,vollmer2014robots}. The degree of contingency of robot feedback also influences scaffolding behavior in complexity and structuring properties \citep{fischer2013impact}.
% Feedback congruent with architecture

% \subsubsection{Approaches to Robot Feedback}

% Action Representation as Feedback 

% Filter as Feedback? "40 years of cognitive architectures: core cognitive abilities and practical applications"
% (attention, restrictions, relevance: Priming based on task demands, domain knowledge, external stimuli, mission context, others such as the field of view.)

% Symbolic vs Geometric

% cleaning task:  "Inferring the Effects of Wiping Motions based on Haptic Perception"

In a study designed to investigate human teacher interpretations of a robot's feedback upon an action demonstration \citep{vollmer2014robots} showed that a robot imitating or emulating an action trajectory is a strong cue to the robot's interpretation of the task as being path- vs goal-oriented. 
\citep{hindemith2021robots} have shown that a configuration interface which is a highly contingent feedback interface can improve the user's mental model about the robot's capabilities.

% Learning
\subsubsection{Robots learning from lay human teachers}
To enable co-construction with human teachers with state-of-the-art machine learning approaches, \citet{vollmer2018user} have shown that a robot can learn a motion skill by relying on lay human feedback instead of a predefined cost-function, and that the robot’s learning performance is better when the learning approach allows the human tutor to provide comparative rather than absolute feedback about the quality of the learned motion \citep{hindemith2021robots}. Yet, teaching in HRI is restricted through given input modalities and it remains mainly within the boundaries of pre-defined interaction protocols that do not allow for scaffolding strategies \citep{vollmer2016pragmatic}.

%ToDo:
%Yukie,
%Andrea Thomaz,
%Maya Cakmak,
%Pierre-Yves Oudeyer: intrinsic curiosity, also as a drive to close knowledge and skill gaps, curriculum learning, zone of proximal development, 

\section{Research Desiderata}

\begin{wrapfigure}{t}{0.3\textwidth}
  \begin{center}
    \includegraphics[width=0.3\textwidth]{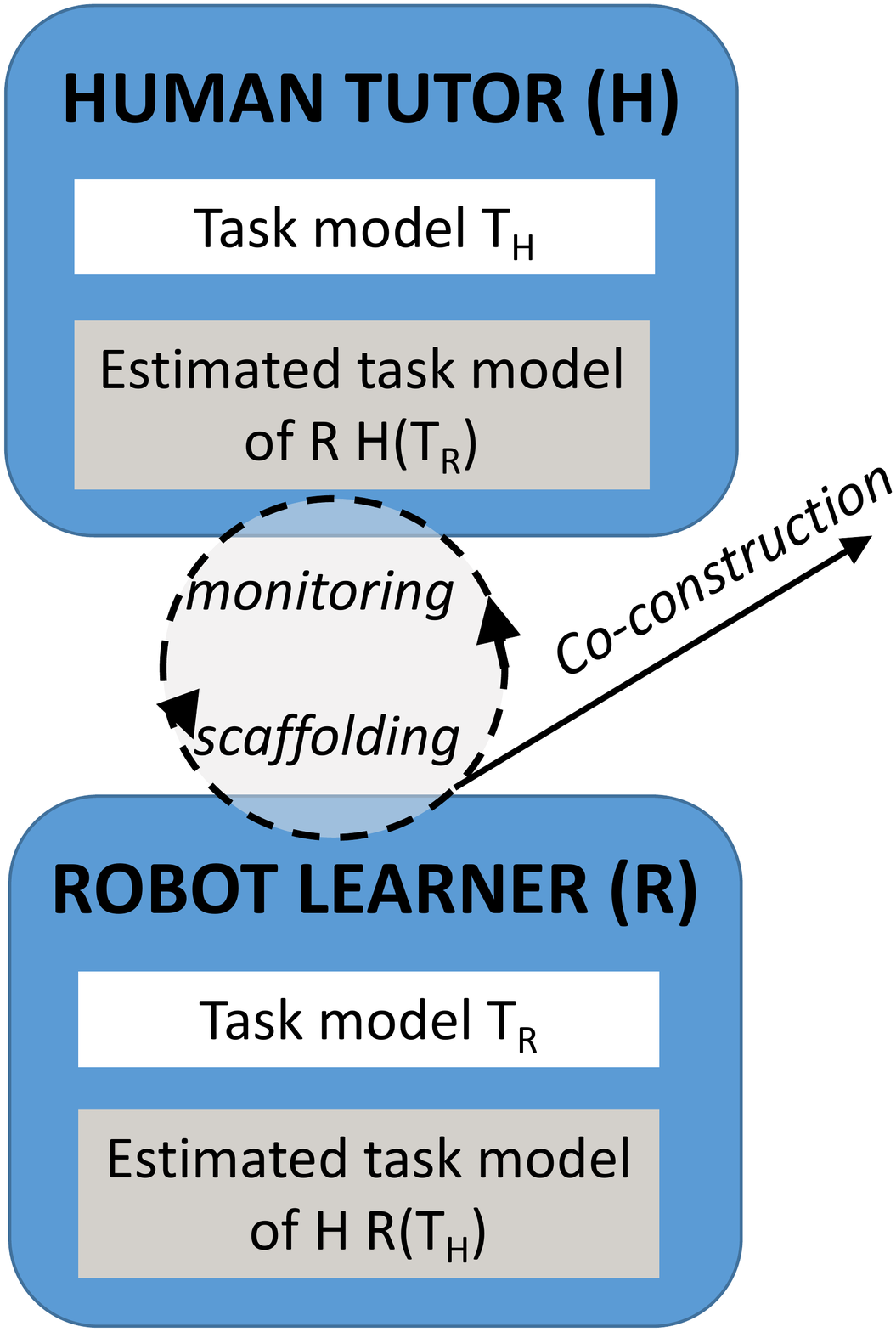}
  \end{center}
  \caption{Co-constructive task-learning process, inspired by the co-construction process proposed by \cite{rohlfing2020explanation}.}
  \label{fig:coco}
\end{wrapfigure}

%Thus, current research seems to converge towards an understanding that the capability for robots to cope with an open ended world requires (1) interaction based on sharable representations and (2) autonomy based on an internal model that supports learning and generalization, be it through intrinsic motivation or through a world model that enables reasoning in new contexts. 
Thus, co-constructive task learning requires a robot to be capable of acquiring and iteratively refining a task model from a human teacher by making use of scaffolding strategies.
Inspired by the co-construction process  for explainable AI as proposed by \cite{rohlfing2020explanation} and as depicted in Fig. \ref{fig:coco} we envision a co-constructive task-learning process as an interaction that requires the human tutor H to provide an explanation and/or demonstration of a task based on her task model ${T_H}$ and to closely monitor the robot learner R's feedback to infer R's Task model ${T_R}$. This inferred task model ${H(T_R)}$ then serves as a basis for H to provide scaffolding in various forms, e.g. by adapting the task so that it can be achieved by R or by providing relevant cues to improve R's task model. This way, the next move of H will contribute to the co-construction of a joint task understanding.

A more detailed roadmap how to address these challenges has yet to be formulated.

\subsection{Scaffolding and Learning Strategies}

We argue that robots for open ended world settings need the ability to be scaffolded in order to learn from humans.
As a first step towards this end we need a better understanding and a systematic overview of the role that different scaffolding strategies play in learning, also in situations where beneficial scaffolding could potentially be elicited by the robot.
In human-robot interaction, we view the following functions of human tutoring as bearing most potential for being beneficial for robotic systems:
(1) Positive or negative feedback and guidance on a task trial, optimally concurrent to provide feedback just in time to enable precise localisation of the source of positive or negative feedback, (2) affective processing to provide implicit cues for the evaluation of different task steps, including motivation, (3) structuring of information: provide meaningful segments with sub-goals that are easier to learn,
(4) hyperpronounced motor execution to provide optimal targets for a reinforcement learning approach that support generalisation to unknown motor configurations.

However, this unstructured list needs to be developed further into a structured map that allows to extract generalised relationships between scaffolding behavior and their effect on learning.

%However, we need a better understanding / systematic overview on the role that different scaffolding strategies play in learning.

%In human-robot interaction, we view the following functions of human tutoring as bearing most potential for being beneficial for robotic systems:

\subsection{Architecture}

How to bring together modules that enable interaction between a human teacher and the robot learner on the one hand and those modules that provide the reasoning capabilities based on sharable representations on the other hand is yet quite unclear. 
For a specific task this problem can be solved by determining explicit rules for specific situations \citep{clodic2007study}. However, solutions that generalise to different situations and tasks, i.e. to an open world, require different architectures.

One of the earliest cognitive architectures is Soar \citep{laird2019soar}. It is based on general cognitive principles that underlie goal-oriented behavior.
One basic assumption is that intelligent behavior that targets to reach a goal can be generated by a search process in a state-based search or problem space. 
%developed as a natural language understanding system 
%the underlying framework has a strong focus on dialog and interaction, emphasizing the need for incremental processing. 
Soar foresees different learning mechanisms for symbolic as well as for procedural knowledge derived from different memory systems.
Accordingly, the modules of the Soar architecture are focused on different memories to which different learning mechanisms are applied.
It thus provides insightful solutions for the coordination of differently represented memories and knowledge sources. 
Note, that this architecture is the basis for the account of Interactive Task Learning \citep{laird2017interactive}. However, interaction has so far received less attention.

The \emph{Cognitive Robot Abstract Machine} (CRAM) \citep{beetz2010cram} is a cognitive architecture that allows to infer through reasoning a plan of concrete motion patterns to realise an underspecified task instruction. It has been continuously improved and extended and now contains additionally an internal simulation component that allows to predict action outcomes. Also, with RoboCog a component has been added for the interpretation of human intentions while acting out a task, thus introducing a kind of partner model \citep{tenorth2009knowrob,beetz2018know,haidu2016action}.

Overall, cognitive architectures focus on the realisation of intelligent autonomous behavior of the robot, but tend to leave out the interplay of components to achieve a meaningful interaction and to integrate a teacher's advice into the own learning process.
An architecture that supports co-constructive learning thus needs to provide mechanisms that can make use of scaffolding strategies for learning. 
%In a new proposal for explainable AI \cite{Rohlfing-XAI} propose the integration of a dynamic partner model to enable the system to provide individually tailored explanations to a user. 

%explainability and transparency to allow teacher to monitor progress and estimate the zone of proximal development

%Vergleich zu Lit

%Vergleich zu Lit

% \subsection{Dialog}

%\subsubsection{Requirements and Approaches}
% I don't recall if this was intended for me (Daniel)
% \citet{sigaud2022towards} call for autonomous, hierarchical few-shot learners.

\subsection{Flexible Representation}
Flexible representations play a critical role in facilitating learning through scaffolding in robotics. 
For example, the selected representations could be required to effectively operate with small sequences before gradually expanding to larger units. Or vice versa a coarse motion could become more structured with more details.
Thus, hierarchical representations are essential in directing attention and expectations towards the appropriate stimuli, enabling efficient learning and the gradual accumulation of knowledge.
To illustrate, consider the process of teaching a child to color a figure, which involves several incremental steps. 
Initially, the child is instructed to crayon only on the sheet of paper before moving on to crayon within the lines of the actual figure. 
To provide such guidance, it is preferable to define constraints in the task space, which is more intuitive than describing limitations in body coordinate systems. %with may involve abstract terms such as left or right. 
In this regard, a constraint-based movement representation based on geometric features, as proposed by \citet{bartels2013constraint}, can be utilized.
The use of such a representation for scaffolding in robotics offers the advantage of designing user interfaces in a natural way.
%, without requiring the user to have prior knowledge of robotics. 
%Consequently, the process of learning new robot skills could become as intuitive as guiding a child to draw.

To render representations more explainable and enable the (re-)definition of goals through natural language, one might train a large language model (LLM) based on prior knowledge that is specially tailored to the problem.
One potential idea here is to leverage knowledge handled in a more structured format, such as Planning Domain Definition Language (PDDL) \citep{fox2003pddl2}.
This would come with the advantage, that the robot would be able to communicate its internal state, otherwise difficult to understand for a human, in natural language while it is less likely that the communicated information contains errors.
Furthermore, otherwise ambiguously formulated goals can be grounded in deterministic knowledge by the human operator.
Together, the robot and the human can this way enter into a dialog to not only discuss current states and events, but also refine prior knowledge, e.g. the effects of actions and thus engage in a co-construction of task knowledge.

\subsection{Interaction Model}

To co-constructively learn a task requires an interaction model that supports crucial features of scaffolding.

Task demonstrations usually involve \emph{multiple modalities} \citep{vollmer2016pragmatic} where speech and other communicative signals accompany the action execution. For an interaction model this means that it needs to analyze multiple modalities in parallel and combine the information from the different modalities according to their semantic and pragmatic content. More specifically, attention guidance through gaze and verbal cues, acoustic packaging through multimodal synchrony for segmentation, and reference to already known tasks are only a few examples of scaffolding cues that should be handled by the interaction model.

Also, the continuous nature of actions allows the human teacher to continuously give feedback to the action demonstration of the robot. This requires incremental processing of both, the perceived communicative signals as well as the action sequences in order to provide a learning component with exact timing information of positive or negative reward for yielding a certain trajectory or goal position.

Importantly, human interactions especially with young learners build on repeating interaction patterns, or \emph{Pragmatic Frames} (PFs). These patterns provide context with respect to the task goal, the means, the roles of the interaction partners etc. \citep{rohlfing2016alternative} which reduces uncertainties and thus supports to learn the meaning of new tasks or related elements.
It is thus important for the interaction component to detect (new) PFs and to build new PFs on the fly in interaction with a human. Current statistical dialog modeling approaches that build on Markov Chains, such as Partially Observable Markov Decision Processes (POMDPs), and that are widely used in Human-Computer and Human-Robot-Interaction do not seem appropriate for the modelling of such recurring patterns because of the restricted history they take into account. 

Especially challenging is the requirement for freedom in scaffolding and the selection of scaffolding cues (i.e., the possibility to jointly construct new or adapt existing PFs) allowing to adapt to new interaction patterns.
This not only requires a flexible interaction protocol but also the capability to detect ostensive (i.e., scaffolding) cues in object manipulation settings.

Overall, this requires an interaction model that is  able to learn and adapt interaction patterns from little data and to integrate context as provided through PFs into the task learning process.

%incremental wrt interaction, 
%flexible wrt representation, 
%needs to be teachable (Freiheit im scaffolding / Eingabemoeglichkeiten, 
%flexible interaction protocol / pragmatic frame,), 
%detection of scaffolding cues / ostension and interactional tension ("Spannungsbogen")

%surface  interaction mit architecture
%fehler, unk knowledge

\subsection{Transparency and Explainability}
In co-construction and scaffolding, the teacher should be able to interpret the mental model of the learner. 
In the case of a robot being trained, this requires robot feedback. Feedback such as eye gaze and queries in active learning can additionally be employed by robots to close knowledge and skill gaps \citep{chao2010transparent, de2015robots}, where intrinsic motivation and curiosity can serve as a driver \citep{oudeyer2007intrinsic}. However, though human teachers can easily interface with robots giving social feedback, like human-like gaze behavior, this has the drawback of eliciting the formation of mental models about the functioning of the robot which are incongruent to how it functions in reality - often resulting in interaction breakdowns and errors due to unmet expectations\footnote{Although, \citet{Wrede2023-comment} argue that breakdowns may have the function of advancing the interaction and may contribute to a better understanding of the robot's inner working.}. Robot feedback, thus, necessarily needs to be born of internal functions as for example as technical transparency mechanisms \citep{hindemith2021robots} or visualizations.
As an interface for the human that is both intuitive and expressive enough to visualise the agent's thought process, attention-driven design, which has been shown to help reduce complexity while maintaining overall situational awareness \citep{schmaus2019knowledge} could be used. 
However, in the case of mismatched assumptions about the state of the world on either side, communication becomes difficult. 
For example, if the robot is unable to perform a task due to unfulfilled preconditions, the human in the loop may not be able to understand why the system is not behaving as expected.
Thus, communication of errors, be they errors in prior knowledge or planning errors, must be built into any human-robot interaction process, especially when it comes to co-construction of task knowledge.

\section{Summary}
We have introduced the idea of co-constructive task learning which extends the notion of interactive task learning towards an even more interactive process -- mutually adaptive, flexible, multimodal. Co-construction, however, gives rise to some requirements concerning the sensitivity and processing of scaffolding behavior, the system architecture and representation of knowledge, which need to be transparent and explainable.

% Zusammenfassung der Abgrenzung zu ITL and others
% schneller lernen durch scaffolding

% Ich habe die Acknowledgments jetzt auskommentiert, weil sie auf der 1. Seite sind (britta)
%\section*{Acknowledgment}
% Die Funding Info soll laut Template auf der ersten Seite in einer Footnote stehen. Put sponsor 
%acknowledgments in the unnumbered footnote on the first page.

%The preferred spelling of the word ``acknowledgment'' in America is without 
%an ``e'' after the ``g''. Avoid the stilted expression ``one of us (R. B. 
%G.) thanks $\ldots$''. Instead, try ``R. B. G. thanks$\ldots$''. Put sponsor 
%acknowledgments in the unnumbered footnote on the first page.

%\section*{References}
\bibliographystyle{IEEEtranN}
\bibliography{conference_101719}

% \begin{thebibliography}{00}
% \bibitem{b1} G. Eason, B. Noble, and I. N. Sneddon, ``On certain integrals of Lipschitz-Hankel type involving products of Bessel functions,'' Phil. Trans. Roy. Soc. London, vol. A247, pp. 529--551, April 1955.
% \bibitem{b2} J. Clerk Maxwell, A Treatise on Electricity and Magnetism, 3rd ed., vol. 2. Oxford: Clarendon, 1892, pp.68--73.
% \bibitem{b3} I. S. Jacobs and C. P. Bean, ``Fine particles, thin films and exchange anisotropy,'' in Magnetism, vol. III, G. T. Rado and H. Suhl, Eds. New York: Academic, 1963, pp. 271--350.
% \bibitem{b4} K. Elissa, ``Title of paper if known,'' unpublished.
% \bibitem{b5} R. Nicole, ``Title of paper with only first word capitalized,'' J. Name Stand. Abbrev., in press.
% \bibitem{b6} Y. Yorozu, M. Hirano, K. Oka, and Y. Tagawa, ``Electron spectroscopy studies on magneto-optical media and plastic substrate interface,'' IEEE Transl. J. Magn. Japan, vol. 2, pp. 740--741, August 1987 [Digests 9th Annual Conf. Magnetics Japan, p. 301, 1982].
% \bibitem{b7} M. Young, The Technical Writer's Handbook. Mill Valley, CA: University Science, 1989.
% \end{thebibliography}
% \vspace{12pt}
% \color{red}
% IEEE conference templates contain guidance text for composing and formatting conference papers. Please ensure that all template text is removed from your conference paper prior to submission to the conference. Failure to remove the template text from your paper may result in your paper not being published.

\end{document}